# CONTOUR DETECTION USING CONTRAST FORMULAS IN THE FRAMEWORK OF LOGARITHMIC MODELS

Vasile PĂTRAŞCU
Department of Informatics Technology
Romanian Air Transport Company (TAROM)
e-mail: vpatrascu@tarom.ro

Vasile BUZULOIU
Image Processing and Analysis Laboratory (LAPI)
Faculty of Electronics and Telecommunications
"Politehnica" University of Bucureşti
e-mail: vbuzuloiu@alpha.imag.pub.ro

*Abstract* – **In this paper we use a new logarithmic model of image representation, developed in [1,2], for edge detection. In fact, in the framework of the new model we obtain the formulas for computing the "contrast of a pixel" and the "contrast" image is just the "contour" or edge image. In our setting the range of values is preserved and the quality of the contour is good for high as well as for low luminosity regions. We present the comparison of our results with the results using classical edge detection operators.**

## I. INTRODUCTION

The contour or edge detection is an essential operation in any chain of image processing aiming at the segmentation of the image. Very often the methods using the classical image representation – i.e. when the pixel value is considered a real number and the addition and scalar multiplication are the classical ones in the field of real numbers – don't end at satisfactory results. This is the case especially for images with regions of low or high luminosity. A correct (edge detection) contouring for these images asks for many intermediate transformations and fitting parameters as well as substantial increase in processing time. An efficient solution to perform contouring is to work in an algebraic structure of logarithmic type.

In this paper we shall concisely present an algebraic structure based on a logarithmic model developed in [1,2]. In this framework we shall define the relative and the absolute contrast between two pixels and the contrast of pixel ( in an arbitrary neighborhood). With these notions we shall obtain an efficient procedure for contour detection: in fact the contour image is the image of the contrast of the pixels.

The remainder of the paper is structured as follows: the section 2 shows the notion of logarithmic function of gray levels which mathematically defines an achromatic image, section 3 shows the vector space of gray levels in the context of logarithmic model, section 4 shows the fundamental isomorphism between the vector space of logarithmic gray levels and the vector spaces of real numbers, section 5 shows the contrast formulas in the logarithmic model, section 6 defines the contour operation in the sense of this paper and shows the contour images using the contrast defined in the logarithmic model elaborated in [1,2] and as a comparation , the same images contoured by classical methods. Finally, section 7 is for short conclusions.

## II. IMAGES AND LOGARITHMIC FUNCTION OF GRAY LEVELS

An image with gray levels (not colors) is defined by a function $l$ with two spatial variables $l(x, y)$,

$l : D \to [0, A]$ where $D \subset R^2$. This function is proportionally with the luminance of image so it takes only finite and positive values. Because we must work with images like vector space elements, the mathematically model almost always used is the $L^2(D)$ space. In this case, we implicitly accept that the values of luminance $l$ can be any real number [3], [4]. The operation of addition and scalar multiplication is the usual operations between real numbers. There are strong reasons (like a physical nature) that make this model a criticisable one because of his inadequation. Otherwise this model is very simple and satisfactory. These aspects guided to the appearance of the representation with others models [5-10]. Oppenheim model [5,6], and Jourlin-Pinoli model [7-10] are logarithmic models; the first using as a set of gray levels the interval $E = [0, \infty)$ and the latter $E = (-\infty, M)$ with $M>0$. Both models have an insufficiency because the set is unboundary. The model showed in [1,2] has $E = (-M, M)$ and $M>0$ to simplify formulas calculating. We called $E = (-M, M)$ the interval of logarithmic gray levels.

This is linked to the "physical space" by a linear function $t : [0, A] \to E$. Thus we obtain the function $f : D \to E$, called the function of logarithmic gray levels that is associated to an image. We will note $F(D, E)$ the set of functions of logarithmic gray levels.

## III. THE REAL VECTOR SPACE OF LOGARITHMIC GRAY LEVELS

The set $E$ of gray levels can be organized like a real vector space, defining an addition (noted by $\langle + \rangle$) and a real scalar multiplication (noted by $\langle \times \rangle$) by the next relations:

$$\forall u, v \in E, \quad u \langle + \rangle v = \frac{u + v}{1 + \frac{u \cdot v}{M^2}} \quad (1)$$

$$\forall \lambda \in R, \forall u \in E,$$
$$\lambda \langle \times \rangle u = M \cdot \frac{(M + u)^\lambda - (M - u)^\lambda}{(M + u)^\lambda + (M - u)^\lambda} \quad (2)$$

We can prove very simple [1], that in this algebraical structure, the opposite of an logarithmic gray level $v \in E$ is $-v$ and therefore the subtraction operation (noted by $\langle - \rangle$) can be defined naturally:

$$\forall u, v \in E, \quad u \langle - \rangle v = \frac{u - v}{1 - \frac{u \cdot v}{M^2}} \quad (3)$$

In the formulas (1), (2), and (3) the operations made in right side with $u, v, M$ values are interpreted as a real numbers operations (usually acceptation).

In natural mode the addition $\langle + \rangle$, scalar multiplication $\langle \times \rangle$ and subtraction $\langle - \rangle$ defined for E can be extended [1,2], also for the set of images $F(D, E)$. We immediately can also prove that $(F(D, E), \langle + \rangle, \langle \times \rangle)$ is a real vector space [1].

## IV. THE FUNDAMENTAL ISOMORPHISM

The real vector spaces $(E, \langle + \rangle, \langle \times \rangle)$ and $(R, +, \times)$ are isomorphous. This isomorphism is defined by the functions:

$\varphi : E \to R, \forall x \in E,$

$$\varphi(x) = \frac{M}{2} \cdot \ln\left(\frac{M + x}{M - x}\right) \quad (4)$$

and his inverse:

$\varphi^{-1} : R \to E, \forall x \in R,$

$$\varphi^{-1}(x) = M \cdot \frac{e^{\frac{x}{M}} - e^{-\frac{x}{M}}}{e^{\frac{x}{M}} + e^{-\frac{x}{M}}} \quad (5)$$

Because the isomorphism $\varphi$ is a logarithmical function we can say that the algebraical structure of gray levels $(E, \langle+\rangle, \langle\times\rangle)$ ground a logarithmic model for image processing.

## V. CONTRAST DEFINITION IN A VECTOR SPACE OF IMAGES $F(D,E)$

The structure of vector space defined for set $F(D,E)$ and the isomorphism defined above allow us to translate the notion of contrast from classical framework (the set of real numbers). We will note $p_i = (x_i, y_i)$, the pairs of coordinates that define the spatial position of a pixel in an image.

### A. The relative contrast between two pixels

The relative contrast between two distinct pixels $p_1, p_2 \in D$, for an image $f : D \to E$ is a logarithmic gray level noted $C_R(p_1, p_2)$, defined by relation:

$\forall p_1, p_2 \in D, p_1 \neq p_2,$

$$C_R(p_1, p_2) = \frac{1}{d(p_1, p_2)} \langle\times\rangle \frac{f(p_1) - f(p_2)}{1 - \frac{f(p_1) \cdot f(p_2)}{M^2}} \quad (6)$$

where $d(p_1, p_2)$ is the Euclidean distance between $p_1$ and $p_2$ in the $R^2$ plain.

### B. The absolute contrast between two pixels

From the relative contrast $C_R$ we can defined the absolute contrast by the formulas:

$\forall p_1, p_2 \in D, p_1 \neq p_2,$
$$C_A(p_1, p_2) = |C_R(p_1, p_2)| \quad (7)$$

### C. The contrast for a pixel

The contrast for an arbitrary pixel $p \in D$, for an image $f \in F(D,E)$ is a positive logarithmic gray levels image, noted $C(p)$, defined by the mean of absolute contrast between the pixel $p$ and the pixels $(p_i)_{i=1,n}$ that belong to a neighborhood $V$. Thus we have the next formula:

$\forall p \in D$,

$$C(p) = \frac{1}{n} \langle\times\rangle \left( \langle\stackrel{n}{+}\rangle_{i=1} C_A(p, p_i) \right) \quad (8)$$

Usually $n=8$ and the neighborhood $V$ is a window with $3 \times 3$ dimensions and heaving the pixel $p$ in the center.

## VI. EXPERIMENTAL RESULTS

Figure 1, 2, 3, 4 show the contouring of images (from [11]) with the contrast formulas defined in section 5. We define $C(p)$ as the contour image for an initial image, which obtained for each pixel using the relation (8) for the contrast calculus. The passing operator from image $f(p)$ to image $C(p)$ is a neighborhood operation that implies only the first order neighborhood. The classical contour operation can be done by many methods and as a comparation we will illustrate with the gradient and the Laplace operators with the next formulas:

$$G_x = \begin{pmatrix} -1 & 0 & 1 \\ -1 & 0 & 1 \\ -1 & 0 & 1 \end{pmatrix},$$

$$G_y = \begin{pmatrix} -1 & -1 & -1 \\ 0 & 0 & 0 \\ 1 & 1 & 1 \end{pmatrix}$$

and

$$L = \begin{pmatrix} 2 & -1 & 2 \\ -1 & -4 & -1 \\ 2 & -1 & 2 \end{pmatrix}$$

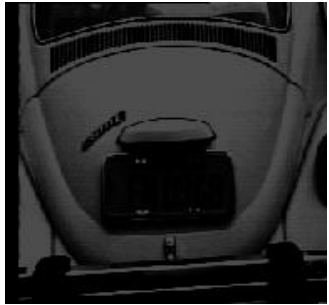

Fig.1 (a) Original image

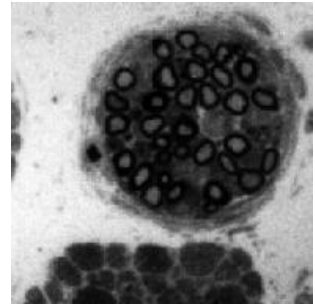

Fig.2 (a) Original image

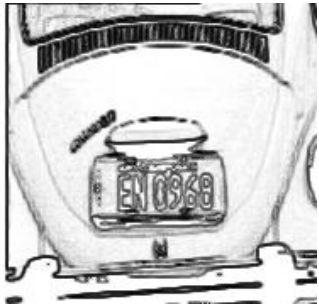

Fig.1 (b) The contrast C(p)

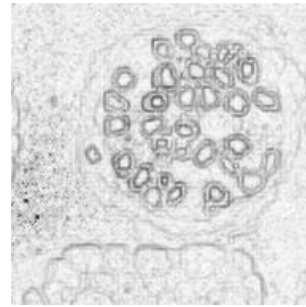

Fig.2 (b) The contrast C(p)

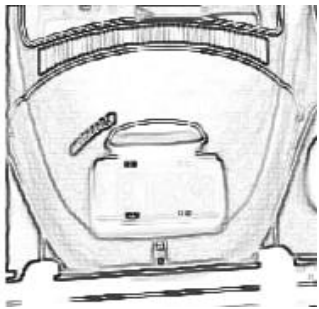

Fig.1 (c) Contour by gradient

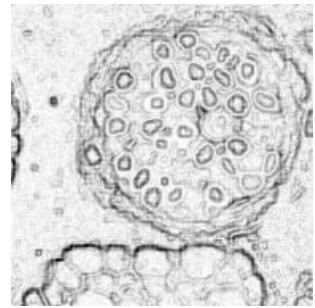

Fig.2 (c) Contour by gradient

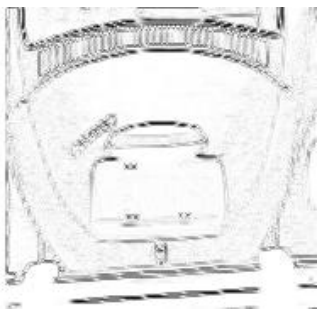

Fig.1 (d) Contour by Laplace operator

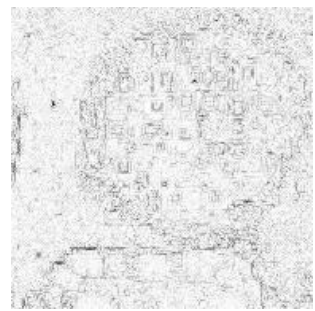

Fig.2 (d) Contour by Laplace operator

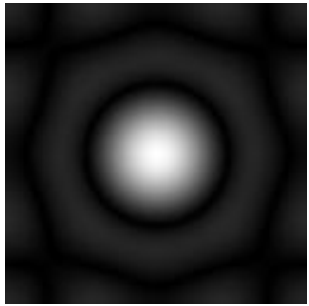

Fig.3 (a) Original image

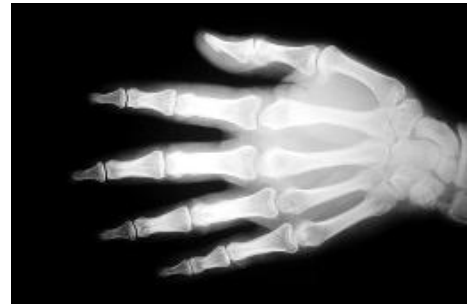

Fig.4 (a) Original image

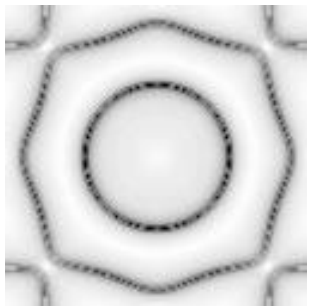

Fig.3 (b) The contrast C(p)

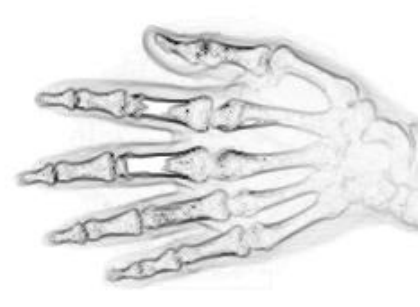

Fig.4 (b) The contrast C(p)

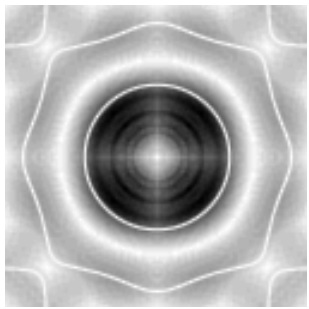

Fig.3 (c) Contour by gradient

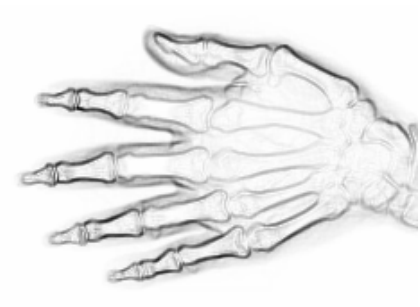

Fig.4 (c) Contour by gradient

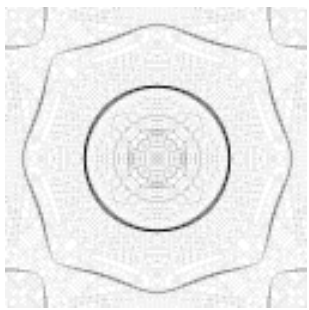

Fig.3 (d) Contour by Laplace operator

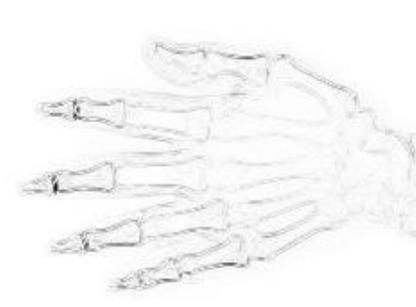

Fig.4 (d) Contour by Laplace operator.

## VII. CONCLUSIONS

The algebraical model of logarithmic type worked out in [1,2] allowed us to define some contrast formulas. These were used in contour operation of images. This contour procedure proves to be more efficient than the classical methods (that is those which use classical operations), as it was shown by all the examples.

The next step involves an establishment of some quantitative evaluation criterions of the contour quality.